\documentclass[sigconf,authorversion]{acmart}
\AtBeginDocument{%
  \providecommand\BibTeX{{%
    \normalfont B\kern-0.5em{\scshape i\kern-0.25em b}\kern-0.8em\TeX}}}
\usepackage{geometry}
\usepackage{tcolorbox}
\usepackage{colortbl}
\usepackage{xcolor}
\usepackage{listings}
\usepackage{amsmath}
\usepackage{algorithmic}
\usepackage{algorithm}
\usepackage{bm}
\usepackage{booktabs}
\usepackage{comment}
\usepackage{multirow}
\usepackage{framed}

\usepackage{url}
\makeatletter
\newcommand{\figcaption}[1]{\def\@captype{figure}\caption{#1}}
\newcommand{\tblcaption}[1]{\def\@captype{table}\caption{#1}}
\makeatother

\newcount\K
\def\bla#1{
\K=0 \loop\ifnum\K<#1
{\textcolor[gray]{0.9}{{\it bla bla bla bla bla bla bla bla bla bla bla bla bla bla bla}}}
\advance\K by1\repeat
}

\newcommand{\todoz}[1]{
\ifx#10
\textcolor{red}{$0.00_{\pm 0.00}$}
\else
\textcolor{red}{#1}
\fi
}

\def\rcode{r_{\text{\scriptsize code}}}
\def\rsp{r_{\text{\scriptsize sp}}}
\def\rpre{r_{\text{\scriptsize pre}}}
\def\ppre{p_\text{\scriptsize pre}}
\def\pprehat{\hat{p}_\text{\scriptsize pre}}
\def\pdef{p_{\text{\scriptsize def}}}
\def\pdefhat{\hat{p}_{\text{\scriptsize def}}}
\def\pinst{p_{\text{\scriptsize inst}}}
\def\pcls{p_{\text{\scriptsize cls}}}
\def\figvspacebottom{\vspace{-12pt}}
\def\tabvspacemid{\vspace{-6pt}}
\def\tabvspacebottom{\vspace{-6pt}}

\begin{document}
\title{AdaCoder: Adaptive Prompt Compression for Programmatic Visual Question Answering}
\author{Mahiro Ukai}
\affiliation{%
\institution{Tokyo Institute of Technology}
\country{Tokyo, Japan}
}
\author{Shuhei Kurita}
\affiliation{%
\institution{National Institute of Informatics}
\country{Tokyo, Japan}
}
\author{Atsushi Hashimoto}
\affiliation{%
\institution{OMRON SINIC X Corporation}
\country{Tokyo, Japan}
}
\author{Yoshitaka Ushiku}
\affiliation{%
\institution{OMRON SINIC X Corporation}
\country{Tokyo, Japan}
}
\author{Nakamasa Inoue}
\affiliation{%
\institution{Tokyo Institute of Technology}
\country{Tokyo, Japan}
}
\renewcommand{\shortauthors}{Ukai, et al.}
\begin{abstract}
Visual question answering aims to provide responses to natural language questions given visual input.
Recently, visual programmatic models (VPMs), which generate executable programs to answer questions through large language models (LLMs), have attracted research interest.
However, they often require long input prompts to provide the LLM with sufficient API usage details to generate relevant code. To address this limitation, we propose AdaCoder, an adaptive prompt compression framework for VPMs.
AdaCoder operates in two phases: a compression phase and an inference phase.
In the compression phase, given a preprompt that describes all API definitions in the Python language with example snippets of code, a set of compressed preprompts is generated, each depending on a specific question type.
In the inference phase, given an input question, AdaCoder predicts the question type and chooses the appropriate corresponding compressed preprompt to generate code to answer the question.
Notably, AdaCoder employs a single frozen LLM and pre-defined prompts, negating the necessity of additional training and maintaining adaptability across different powerful black-box LLMs such as GPT and Claude.
In experiments, we apply AdaCoder to ViperGPT and demonstrate that it reduces token length by 71.1\%, while maintaining or even improving the performance of visual question answering.
\end{abstract}
\maketitle
\begin{figure}
\vspace{4pt}
\centering
\includegraphics[width=0.98\linewidth]{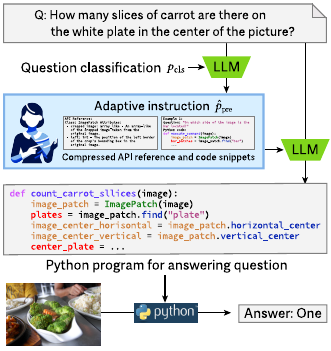}
\vspace{-12pt}
\caption{Inference procedure of AdaCoder. Given an input question, an adaptive programming instruction $\pprehat$ for the specific question type is used to generate a Python program for visual question answering.}
\label{fig:teaser}
\vspace{-14pt}
\end{figure}

\section{Introduction}
Visual question answering (VQA), which aims to automatically provide answers to questions related to visual content, is a challenging research topic in the fields of multimedia analysis, computer vision, and natural language processing~\cite{chen2023vtqa2023, yang2022avqa, antol2015vqa, goyal2017vqav2, suhr2019nlvr2, Hudson2019GQADataset}.
Thanks to the advantages of deep learning techniques, significant progress has been made in VQA over the past decade with end-to-end learning models such as GLIP~\cite{Li2022GLIP}.
However, these models do not explicitly distinguish between visual processing and reasoning, which limits their generalizability and interpretability.

To overcome this limitation, several pioneering studies have introduced visual programmatic models (VPMs), models that generate executable programs specifically designed to answer questions, providing a more manageable and transparent inference process~\cite{Suris2023ViperGPT, Gupta2022VisProg, Subramanian2023CodeVQA, shen2024pyramid, Hu2024VPD, ge2024recursive}.
VPMs typically consist of a large language model (LLM) for code generation and a set of APIs for image processing. Given an input question, the LLM analyzes the text to understand the intent and the required computational steps. It then generates a program that, when executed, can manipulate and analyze images by using APIs to produce the desired answer, where the APIs include both low-level modules ({\it e.g.}, image cropping) and high-level modules ({\it e.g.}, object detection). VPMs have proven effective and are gaining traction; however, they also face challenges in terms of computational complexity, as long prompts are required to enable the LLM to understand API usage for generating appropriate programs.

To reduce computational costs, the development of efficient neural network architectures has been extensively studied. However, these approaches require retraining or additional learning, which is not feasible for application to LLMs trained with huge data, such as GPT and Claude, which we refer to as black-box LLMs. Recently, research has begun to focus on prompt compression~\cite{mu2023learning, jiang2023llmlingua, chevalier2023adapting}, which involves optimizing input prompts to achieve high performance with shorter inputs. For example, LLMLingua~\cite{jiang2023llmlingua, wu2024llmlingua2} compresses prompts using smaller models before using black-box LLMs.

Inspired by these studies, we introduce AdaCoder, a framework of adaptive prompt compression for VPMs. More specifically, AdaCoder operates in two phases: a compression phase and an inference phase.
The compression phase generates a set of compressed preprompts, each depending on a specific question type, given a preprompt that describes all API definitions in the Python language with example snippets of code.

The inference phase adaptively selects a compressed prompt by classifying the question type and generates a Python program to answer the input question, as shown in Figure~\ref{fig:teaser}.
Notably, we implement all of the modules of AdaCoder with a single frozen LLM, which allows implementation with black-box LLMs.
Our contributions are summarized as follows:

{
\setlength{\leftmargini}{16pt}
\begin{enumerate}
\setlength{\itemsep}{2pt}
\item[1)]
We propose AdaCoder, a novel prompt compression framework for VPMs. It adaptively selects a short instruction for code generation based on question type.
\item[2)]
We define and formulate all procedures of AdaCoder with a single frozen LLM. This avoids additional training and enables implementation with black-box LLMs.
\item[3)]
We demonstrate the effectiveness of AdaCoder over the state-of-the-art ViperGPT~\cite{Suris2023ViperGPT} model on three VQA datasets with GPT and Claude.
We show that the token length of input prompts is reduced by 71.1\%, while maintaining or even improving question answering performance. We also show that AdaCoder outperforms LLMLingua~\cite{jiang2023llmlingua} in our evaluations.
\end{enumerate}
}
\section{Related work}
\subsection{Visual question answering}

\noindent \textbf{End-to-end models.}
In the early phase of VQA research history, a number of neural network architectures designed to process multimodal inputs were introduced. These include a combination of a convolutional neural network (CNN) for visual feature extraction and a recurrent neural network (RNN) for textual feature extraction~\cite{malinowski2015ask, huang2021diagnostic, zhang2022context}. Recent models often include attention modules to enhance individual feature extraction for each modality and to combine features of multiple modalities effectively~\cite{yang2016stackedattention, anderson2018bottomupandtopdown, sarkar2022grad, wu2022ques}.
Large-scale pre-training has become a critical component in improving the performance of these models, enabling them to answer complex questions by implicitly associating words with specific regions in images~\cite{li2021unsupervised, le2021vttransformer, Li2022GLIP}.
More recently, LLMs have been incorporated into VQA frameworks with prompt tuning techniques such as self-prompt tuning~\cite{yuan23selfpt}.
However, these models do not explicitly distinguish between visual
processing and reasoning, limiting their interpretability.
Some recent studies have focused on techniques to improve interpretability such as causal inference~\cite{chen23kecvqg}, reasoning path~\cite{Li22aivqa}, reasoning prompts~\cite{lan23questionprompts} and gradient-based explainability method~\cite{wang22towards}.

\noindent \textbf{Visual programmatic models.}
To improve interpretability and generalizability, VPMs that generate programs to answer questions based on visual input have been gaining research attention. This is a novel approach that leads to more manageable and traceable inferences because the generated programs contain logical sequences that are understandable to humans and articulate a step-by-step methodology for reaching conclusions.
Examples of VPMs include ViperGPT~\cite{Suris2023ViperGPT}, VisProg~\cite{Gupta2022VisProg}, and CodeVQA~\cite{roziere2023codellama}.
All of these generate Python programs utilizing image processing APIs, such as object detection, through a frozen LLM. However, generating programs to answer complex and compositional questions requires many APIs and example codes for them. As a result, the length of the input prompt becomes long. To the best of our knowledge, this work is the first to propose adaptive prompt compression for VPMs.

\begin{figure*}
\centering \includegraphics[width=\linewidth]{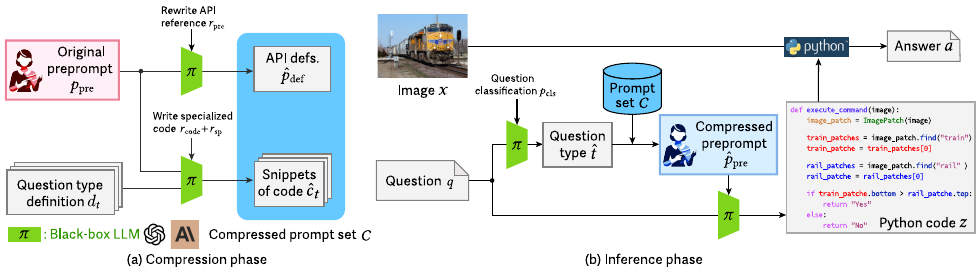}
\caption{AdaCoder framework. (a) Compression phase generates a set of compressed prompts $\mathcal{C}$ by utilizing an LLM $\pi$ with two instructions $\rpre$ for rewriting API definitions and $\rcode + \rsp$ for writing snippets of code specialized for each question type $t$.
(b) Inference phase adaptively selects code snippets to create compressed preprompt $\pprehat$ for generating a Python code $z$ for visual question answering.}
\label{fig:overview}
\end{figure*}

\subsection{Large language models}
\noindent \textbf{Code generation.}
Extensive research and development in the field of natural language processing (NLP) has led to the creation of LLMs that excel at a variety of NLP tasks. Among these, a distinct group of LLMs is specifically designed for programming code generation, having been trained on large amounts of programs and documents related to programming. 
For example, Codex~\cite{Chen2021Codex}, a variant of the GPT-3 lineup, demonstrates its proficiency in multiple programming languages. CodeLlama~\cite{roziere2023codellama}, which is built on Llama2~\cite{touvron2023llama} and has an expanded code dataset, shows improved performance in handling larger contexts in programming.

Most recently, black-box LLMs such as GPT-3.5/4~\cite{brown2020gpt3}, Claude\cite{claude} and Gemini~\cite{team2023gemini} integrate extensive knowledge from a broad spectrum of domains, including programming, allowing them not only to generate code but also to understand and execute complex instructions given by humans.
Since their zero-shot performance on programming tasks is remarkably high, they are expected to automate many aspects of coding in future that were previously manual and time-consuming, and are also useful for integration into VPMs for visual question answering. 

\noindent \textbf{Reasoning and interpretability.}
Interpretability is an important consideration when integrating LLMs into real-world systems, especially in contexts that require high reliability and accountability. Various prompting techniques have significantly improved the interpretability of LLMs. For example, chain-of-thought prompting~\cite{Jin2023TabularCoT}, which provides an LLM with a series of contextual examples, enables intermediate reasoning to reach final conclusions.
Tree-of-thought prompting~\cite{Yao2023TreeOfThoughts_ToT} constructs a tree structure of thoughts, enriching the decision-making process by branching out various reasoning pathways.
VPMs can also be viewed as an extended prompting method that improves interpretability because they show a sequence of logical steps leading to a conclusion by understandable programs.
However, these methods also increase the complexity of input prompts because the instructions for LLMs need to be detailed, thus increasing computational costs.

\noindent \textbf{Prompt compression.} Several strategies have been developed to compress prompts, notably by creating specialized tokens through prompt-based fine-tuning of LLMs~\cite{mu2023learning, chevalier2023adapting, wingate2022prompt, ge2022extensible}, with the goal of minimizing the number of tokens processed during inference. However, fine-tuning of LLMs often limits their generalizability and is not always applicable to black-box LLMs.
Other efforts have focused on token reduction.
These include token pruning during inference~\cite{goyal20powerbert, kim21lengthadaptive, kim22learned} and token merging ~\cite{bolya23tokenmerging}.
However, these methods are generally proposed for small models such as BERT and ViT, and rely on fine-tuning or intermediate inference results. 
Most recently, Jiang \textit{et al.}~\cite{jiang2023llmlingua} have introduced LLMLingua, which compresses prompts with a small model and feeds the compressed prompts to an LLM.
This method can be applied to black-box LLMs because it does not require comprehensive fine-tuning of LLMs.

In contrast to these previous studies, this work aims to define and formulate all procedures of prompt compression and inference for code generation with a single frozen LLM to fully leverage the advantages of powerful black-box LLMs.

\section{AdaCoder framework}
This section introduces AdaCoder, a framework for adaptive prompt compression for VPMs.
Figure~\ref{fig:overview} shows an overview of the AdaCoder framework, which consists of two phases: the compression phase and the inference phase.
The compression phase is run only once to prepare compressed prompts, each of which is specialized for a specific question type.
The inference phase classifies question type and adaptively selects a compressed preprompt to generate code for visual question answering.
Below, we begin with a preliminary formulation of a VPM. We then present each phase of AdaCoder.

\subsection{Preliminary}
\noindent \textbf{Notation and settings.}
We follow the notation used in previous work on VPMs~\cite{Suris2023ViperGPT, Subramanian2023CodeVQA}.
Let $x \in X$ be an input image and $q \in Q$ be an input question about the image, where $X$ is a set of images and $Q$ is a set of questions.
VPMs aim to generate a code $z \in Z$ that returns the answer $a \in A$ to the question, where $Z$ is a set of executable codes and $A$ is a set of answers.

The process of answering questions is divided into two steps: the code generation step and the execution step.
The former generates a code as
\begin{align}
z = \Pi(q),
\end{align}
where $\Pi: Q \to Z$ is a code generation module. The latter executes the code with an input image by
\begin{align}
a = \Lambda(x, z),
\end{align}
where $\Lambda: X \times Z \to A$ is the execution engine. This work utilizes the Python execution engine for $\Lambda$.

\noindent \textbf{Large language model.}
To implement the code generation module, a single frozen LLM $\pi: T \to T$ is often used, where $T$ is a set of texts\footnote{This work assumes that questions, answers, and codes are in text form, {\it i.e.}, $Q, A$, and $Z$ are subsets of $T$.}.
For example, the code generation module $\Pi$ can be defined~by
\begin{align}
\label{eq:codegen}
\Pi(q) = \pi(\ppre + q),
\end{align}
where $\ppre \in T$ is a preprompt that gives instructions to generate code using image and text processing APIs, $q \in Q$ is an input question, and $+$ indicates textual concatenation.
Here, APIs include both low-level functions, such as image cropping, and high-level functions, such as object detection.

\noindent \textbf{Preprompt definition.}
In order to provide the LLM with detailed instructions on how to use the APIs, the preprompt $\ppre$ typically includes API definitions, coding instructions, and example snippets of code.
We define a preprompt $\ppre$ by
\begin{align}
\label{eq:psi}
\ppre = \Psi (\pdef, c, \pinst),
\end{align}
where $\pdef$ is a text of API definitions,
$c \in Z$ is textually concatenated example snippets of Python code,
$\pinst \in T$ is a coding instruction written in a natural language,
and $\Psi$ is a structural aggregation function to insert code snippets to immediately after function definitions as comments.
For example, a code snippet for comparing the positions of objects is inserted immediately after the definition of the object detection function.
Below, we review the preprompt of 
ViperGPT~\cite{Suris2023ViperGPT}, which we use in Section~\ref{sec:experiments}.

\noindent \textit{1) API definitions.}
The text of API definitions for $\pdef \in Z$ is written in Python and includes both class, method and function definitions. Specifically, it consists of the Python class \texttt{ImagePatch} to represent an image patch and a set of auxiliary functions.

\noindent \textit{2) Code snippets.}
For each function and method, one or two code snippets are provided. Each code snippet calls the function or method at least once. For example, the following code snippet is given for the \texttt{find} method that detects objects in images.
\vspace{-4pt}
{
\renewcommand{\baselinestretch}{0.85}
\begin{lstlisting}
# Return the foo
def execute_command(image) -> List[ImagePatch]:
    image_patch = ImagePatch(image)
    foo_patches = image_patch.find("foo")
    return foo_patches
\end{lstlisting}
}

\noindent \textit{3) Coding instruction.}
The coding instruction provides short instructions describing how to write code, specifying a programming language and how APIs should be used.
Specifically, $\pinst$ is the following text:
\vspace{-4pt}
{
\renewcommand{\baselinestretch}{0.85}
\begin{lstlisting}
Write a function using Python and the ImagePatch class
(above) that could be executed to provide an answer to
the query. 

Consider the following guidelines:
- Use base Python (comparison, sorting) for basic logical
  operations, left/right/up/down, math, etc.
- Use the llm_query function to access external
  information and answer informational questions not
  concerning the image.
\end{lstlisting}
}

\noindent \textbf{Token length.}
One of major limitations of previous VPMs is that the input token length is long, resulting in a large computational load.
This work addresses this limitation by introducing an adaptive prompt compression method.
More specifically, we define the input token length of the code generation module in Eq.~(\ref{eq:codegen}) as
\begin{align}
\ell(q; \Pi) = |\ppre| + |q|,
\end{align}
where $|t|$ is the number of tokens of a text $t \in T$. Our goal is to reduce this length.

\subsection{Compression phase}

As shown in Figure~\ref{fig:overview}a, the compression phase creates a set of compressed prompts $\mathcal{C} = \{\pdefhat \} \cup \{\hat{c}_{t}: t \in Y\}$, where $\pdefhat$ is a compressed text of API definitions, $\hat{c}_{t}$ is a compressed code snippet for question type $t \in Y$, and $Y$ is a set of question types. Below, we detail the two-step process for compressing API definitions and code snippets.

\begin{figure}
\centering
\includegraphics[width=\linewidth]{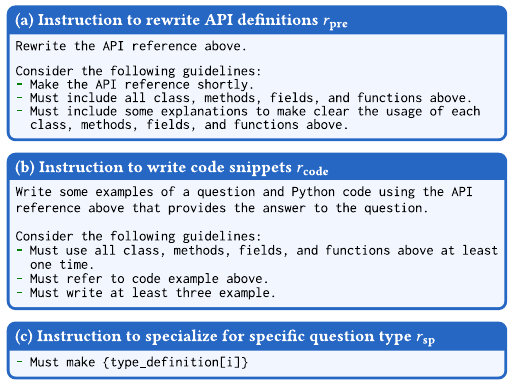}
\caption{Instruction prompts for the compression phase. 
\textnormal{\texttt{\{type\_definition[i]\}}} is a placeholder to which a question type definition $d_{t}$. See Table~\ref{tab:gqa_qtypes} for example type definitions.}
\label{fig:prompt_compress}
\end{figure}

\noindent \textbf{Compressing API definitions.}
This step compresses the API definitions by 
\begin{align}
\label{eq:pdefhat}
\pdefhat = \pi (\ppre + \rpre)
\end{align}
where $\ppre$ is the original preprompt in Eq.~(\ref{eq:psi}), $\pi$ is a frozen LLM, and $\rpre$ is the instruction to rewrite API definitions. Figure~\ref{fig:prompt_compress}a shows the definition of $\rpre$.

\noindent \textbf{Compressing code snippets.}
This step compresses the code snippets for each question type as follows:
\begin{align}
\label{eq:compress_code_snippets}
\hat{c}_{t} = \pi (\ppre + \rcode + \rsp(d_{t})),
\end{align}
where $\ppre$ is the original preprompt,
$\rcode$ is the instruction to write code snippets, $\rsp$ is an additional instruction to write code specialized for a specific question type with a placeholder to insert the definition of question type $d_{t}$, and $t \in Y$ is a question type.
Figure~\ref{fig:prompt_compress}b and \ref{fig:prompt_compress}c show the definitions of $\rcode$ and $\rsp$, respectively.
Here, we assumed that a pre-defined set of question types $Y$ is given.
For example, with the GQA dataset~\cite{Hudson2019GQADataset}, five question types shown in Table~\ref{tab:gqa_qtypes} are provided with their definitions.

\subsection{Inference phase}

The inference phase generates a code to answer the input question by utilizing a compressed preprompt adaptively selected based on the question type as shown in Figure~\ref{fig:overview}b.
More specifically, this phase consists of four steps: question classification, preprompt generation, code generation, and execution.

\definecolor{codeboxcolor}{RGB}{39,103,196}
\definecolor{codeboxcolorl}{RGB}{212,225,250}
\newtcolorbox{codebox}[1]{colback=codeboxcolorl!30!white,colframe=codeboxcolor,fonttitle=\bfseries,title=#1,left=0mm,right=0mm,top=0mm,bottom=0mm,toptitle=-0.2mm,bottomtitle=-0.8mm}

\begin{figure}
\centering
\includegraphics[width=\linewidth]{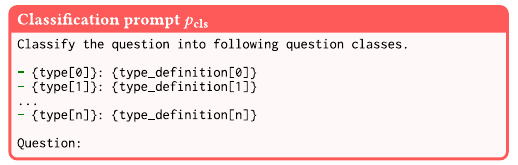}
\caption{Classification prompt for the inference phase. \textnormal{\texttt{\{type[i]\}}} and \textnormal{\texttt{\{type\_definition[i]\}}} are placeholders for names and definitions of question type, respectively, for $i = 0, 1, \cdots, n$ where $n$ is the number of question types.}
\label{fig:prompt_inference}
\end{figure}

\noindent \textbf{Question classification.}
Given an input question $q \in Q$, this step predicts the question type.
We define classification prompt $\pcls$ and use the LLM $\pi$ for question classification as follows:
\begin{align}
\label{eq:that}
\hat{t} = \pi (\pcls + q),
\end{align}
where $\hat{t}$ is the predicted question type.
The classification prompt consists of a short instruction for classification and a list of definitions of question types.
The definition of classification prompt is shown in Figure~\ref{fig:prompt_inference}.

\noindent \textbf{Preprompt generation.}
This step generates a compressed preprompt given the question type as follows:
\begin{align}
\label{eq:pprehat}
\pprehat = \Psi (\pdefhat, p_{\text{\scriptsize inst}}, \hat{c}_{\hat{t}})
\end{align}
where $\pdefhat \in \mathcal{C}$ is the compressed API definitions in Eq.~\ref{eq:pdefhat},
$p_{\text{\scriptsize inst}}$ is the coding instruction, $\hat{c}_{\hat{t}}  \in \mathcal{C}$ is the snippets of code for the question type $\hat{t}$, and $\Psi$ is the structural aggregation function in Eq.~(\ref{eq:psi}).
Note that the computational cost of this step is almost negligible because both compressed prompts, $\pdefhat$ and $\hat{c}_{\hat{t}}$, have already been computed in the compression phase.
We do not compress the coding instruction $\pinst$, because it is already short.

\noindent \textbf{Code generation.}
This step generates a Python code $z$ as follows:
\begin{align}
\label{eq:zhat}
z = \pi ( \pprehat + q),
\end{align}
where $\pprehat$ is the compressed preprompt.

\noindent \textbf{Execution.}
Finally, the predicted answer $a$ to the question is obtained by executing the code as follows:
\begin{align}
a = \Lambda(x, z),
\end{align}
where $x$ is an input image, and $\Lambda$ is the Python execution engine.

\subsection{Discussion}

\noindent \textbf{AdaCoder formulation.}
By substituting Eqs.~(\ref{eq:that}) and (\ref{eq:pprehat}) into Eq.~(\ref{eq:zhat}), we can finally define the code generation module $\Pi_{\text{\scriptsize Ada}}$ of AdaCoder as follows:
\begin{align}
\label{eq:adagen}
\Pi_{\text{\scriptsize Ada}}(q)
= \pi \left(\Psi \left(\pdefhat, \pinst, \hat{c}_{\pi(\pcls + q)}\right) + q \right),
\end{align}
by which a code is generated as $z = \Pi_{\text{\scriptsize Ada}}(q)$.
The total token length is given by
\begin{align}
\ell(q; \Pi_{\text{\scriptsize Ada}}) = 
|\pdefhat| + |\pinst| + |\pcls| +  |\hat{c}_{\pi(\pcls + q)}| + 2 |q|.
\end{align}
Below, we discuss the computational cost and adaptiveness.

\noindent \textbf{Computational cost.}
Although, in the first sight, AdaCoder seems computationally more expensive than the conventional code generation module in Eq.~(\ref{eq:codegen}) because the LLM $\pi$ is called twice in Eq.~(\ref{eq:adagen}); indeed, AdaCoder improves the computational efficiency in practice when a black-box LLM such as GPT or Claude is used for $\pi$ with state-of-the-art VPMs such as ViperGPT, because the input token length is significantly shortened.
Compared to previous prompt compression methods such as LLMLingua, our approach is more efficient and effective because it can
reduce the token length while preserving the structure of code.
We will experimentally demonstrate this in Section~\ref{sec:res}.1.
\newcommand{\methodname}[1]{{\ttfamily\bfseries\textcolor[rgb]{0.364705882,0.215686275,1.0}{#1}}}
\newcommand{\classname}[1]{{\ttfamily\bfseries\textcolor[rgb]{0.0901,0.5020,0.0941}{#1}}}
\newcommand{\varname}[1]{{\ttfamily\bfseries\textcolor[rgb]{0.70,0,0}{#1}}}
\newcommand{\others}[1]{{\ttfamily\bfseries\textcolor[rgb]{0,0,0}{#1}}}
\newcommand{\defs}[1]{{\ttfamily\bfseries\textcolor[rgb]{0.835294118,0.345098039,1.0}{#1}}}

\newcommand{\coms}[1]{{\ttfamily\textcolor[rgb]{0.71372549,0.305882353,0.556862745}{#1}}}

\newcommand{\intname}[1]{{\ttfamily\bfseries\textcolor[rgb]{0.490196078,0.48627451,0.650980392}{#1}}}
\newcommand{\nonename}[1]{{\ttfamily\bfseries\textcolor[rgb]{0.2,0.631372549,0.631372549}{#1}}}

\begin{table}
\centering
\caption{Question type definition for the GQA dataset.}
\label{tab:gqa_qtypes}
\footnotesize
\begin{tabular}{p{0.086\linewidth}p{0.85\linewidth}}
\toprule
Type \(t\) & Definition \(d_{t}\)\\
\midrule
obj & \cellcolor{gray!15}\texttt{question asking existence of object.}\\ \addlinespace[2.5pt]
cat & \cellcolor{gray!15}\texttt{question related to object identification within some category.}\\ \addlinespace[2.5pt]
attr & \cellcolor{gray!15}\texttt{question asking about the attributes or position of an object. (e.g. "What is the color of bar?", "On which of image is the foo?")}\\ \addlinespace[2.5pt]
rel & \cellcolor{gray!15}\texttt{question derived from an affirmative sentence and asking about its subject or object (e.g. "What is the foo next to the baz wearing?", "Is the qux holding a quux?").}\\ \addlinespace[2.5pt]
global & \cellcolor{gray!15}\texttt{question asking about the entire situation of the scene, such as weather or facility (e.g. "Is it foo?").}\\
\bottomrule
\end{tabular}
\end{table}

\noindent \textbf{Adaptiveness.} A major strength of AdaCoder is that it does not require additional training to adaptively compress the preprompt. Since recent black-box LLMs exhibit quite high zero-shot performance on text processing tasks such as text classification and summarization, AdaCoder leverages these capabilities to enhance efficiency and reduce the computational costs of VPMs. 

\section{Experiments}
\label{sec:experiments}

\subsection{Experimental settings}

\noindent \textbf{Datasets.}
We use three VQA datasets for evaluation:
GQA~\cite{Hudson2019GQADataset}, VQAv2~\cite{goyal2017vqav2}, and NLVR2~\cite{suhr2019nlvr2}.
The GQA dataset is designed to test a model's visual reasoning abilities, encompassing five question types: existence of objects (obj), category of objects (cat), attributes of objects (attr), relationships between subjects and/or objects (rel), and global questions (global).
The VQAv2 dataset contains open-ended questions about images that require an understanding of visual content to generate answers.
The NLVR2 dataset is designed to test a model's ability to understand complex natural language statements and their correspondence to a given pair of images.
From each of these two dataset, we randomly choose 2,000 QAs\footnote{This is due to the usage limits of Claude and GPT. The list of sampled QA IDs will be provided along with our code.}.

\begin{table*}[t]
\centering
\setlength{\tabcolsep}{6pt}
\caption{Comparison with other methods.
AdaCoder is compared with ViperGPT~\cite{Suris2023ViperGPT}, LLMLingua~\cite{jiang2023llmlingua}, and Simple compression that omits QA classification prompts.
}
\tabvspacemid
\label{tab:sota}
\begin{tabular}{l|c|ccc|ccc|c}
\toprule
\multirow{2}{*}{Method} & \multirow{2}{*}{LLM} & 
\multicolumn{3}{c|}{Accuracy (\%)} &
\multicolumn{3}{c|}{Input prompt} & Output
\\
&& GQA & VQAv2 & NLVR2 & Token length $\downarrow$ & Characters $\downarrow$ & Reduction $\uparrow$ & Token length \\
\midrule
ViperGPT baseline & gpt-3.5-turbo & 41.3 & 42.7 & 59.2 & 3,434 & 15,950 & - & 78 \\
LLMLingua & gpt-3.5-turbo & 39.1 & 45.2 & 47.3 & 2,536 & 11,507 & 26.2\% & 71 \\
Simple compression & gpt-3.5-turbo & 28.9 & 42.6 & 50.3 & 810 & 3,553 & 76.4\% & 80 \\
AdaCoder (Ours) & gpt-3.5-turbo & \textbf{43.6} & \textbf{46.2} & \textbf{60.8} & 993 & 4,343 & 71.1\% & 77\\
\midrule
ViperGPT baseline & claude-3-haiku & 40.4 & 42.6 & \textbf{60.1} & 3,777 & 15,950 & - & 300\\
LLMLingua & claude-3-haiku & 37.0 & 43.1 & 59.5 & 2,766 & 11,507 & 26.8\% & 306 \\
Simple compression & claude-3-haiku & 14.5 & 23.6 & 54,3 & 1,181 & 4,535 & 68.7\% & 245\\
AdaCoder (Ours) & claude-3-haiku  & \textbf{41.6} & \textbf{44.7} & \textbf{60.1} & 1,170 & 4,503 & 69.0\% & 234\\
\bottomrule
\end{tabular}
\end{table*}

\noindent \textbf{Evaluation metrics.}
We use the exact match accuracy (\%) for case-insensitive answers as a QA performance evaluation metric.
The reduction rate (\%) of the input token length is used to evaluate the compression performance.

\noindent \textbf{Baselines.}
The baselines are ViperGPT~\cite{Suris2023ViperGPT} and LLMLingua~\cite{jiang2023llmlingua} applied to it. They are state-of-the-art VPM and prompt compression method, respectively.

\noindent \textbf{Implementation details.}
We implement AdaCoder on top of the official implementation of ViperGPT\footnote{https://github.com/cvlab-columbia/viper}.
The API set consists of basic image and text processing functions. Specifically, it consists of the \texttt{ImagePatch} class and a set of auxiliary functions. The \texttt{ImagePatch} class is a class to store a image region and has the following nine methods.
{
\setlength{\topsep}{-1pt}
\setlength{\FrameSep}{3pt}
\setlength{\parindent}{0pt}
\begin{framed}
\small
\noindent 1) {crop}, 2) {overlaps\_with}, 3) {find}, 4) {exists}, 5) {best\_text\_match},\\
6) {verify\_property},
7) {simple\_query}, 8) {llm\_query}, 9) {compute\_depth}.
\end{framed}
The auxiliary function set consists of the following four functions.
\begin{framed}
\small
1) distance, 2) best\_image\_match, 3) bool\_to\_yesno, 4) coerce\_to\_numeric.
\end{framed}
Each method or function is provided with its definition in Python and example code snippets.
}

\noindent \textbf{LLMs.}
We use GPT and Claude for both ViperGPT and AdaCoder.
For GPT, we use \texttt{gpt-3.5-turbo}, released as version 1106, which is trained on data up to September 2021 and is provided by the OpenAI platform.
For Claude, we use \texttt{claude-3-haiku}, released as version 20240307. This is a model trained using large amounts of feedback on long document tasks. Note that the original ViperGPT used \texttt{code-davinci-002} (the GPT-3 Codex~\cite{Chen2021Codex}), which was fine-tuned for code generation tasks and is no longer accessible.

\subsection{Experimental results}
\label{sec:res}

\vspace{3pt}
\noindent \textbf{\large \ref{sec:res}.1 \hspace{4pt} Main results}
\vspace{3pt}

\noindent \textbf{QA accuracy.}
Table~\ref{tab:sota} shows QA accuracy in comparison to the ViperGPT baseline. We see that AdaCoder reduces the input token length by 71.1\%, while improving QA accuracy on all of the three datasets. This shows the effectiveness and efficiency of the proposed prompt compression method.

The simple compression setup omits the instruction prompt $\rsp$ to compress for specific question type ({\textit{i.e., }} question type classification is omitted).
We see that the QA accuracy is significant degraded by this omission.

With LLMLingua, we observed that it cannot maintain the structure of code snippets in the preprompt after compression at a reduction rate of 71.1\% (the same rate as ours), resulting in a QA accuracy of 0\%. Therefore, the results in Table~\ref{tab:sota} are given at a lower reduction rate $\simeq$ 25\% by adjusting the compression ratio parameter accordingly.
With this setting, LLMs can generate executable code with a probability of 98\%; however, the QA accuracy is degraded by 2.2 points on GQA.
This shows that prompt compression for VPMs is challenging.
It is also worth noting that while our method is effective, it relies on QA classification and is not inherently a robust compression approach across all problems.

\begin{figure}
\centering
\includegraphics[width=\linewidth]{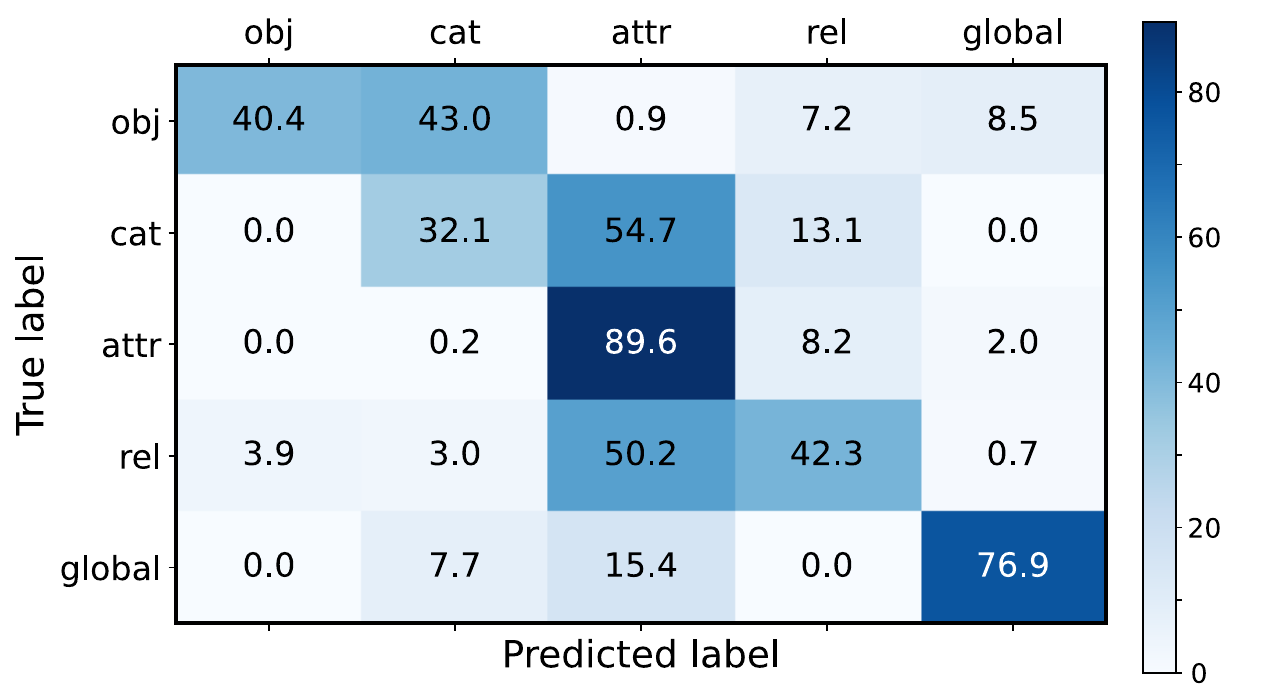}
\caption{Confusion matrix of question classification. Overall classification accuracy was 58.1\%. (GQA dataset, gpt-3.5-turbo)}
\label{fig:confusion}
\figvspacebottom
\end{figure}

\noindent \textbf{Question type classification.}
Figure~\ref{fig:confusion} shows the confusion matrix of question type classification for the GQA dataset. We observe that two question types, ``attr'' and ``global'', achieve accuracies greater than 75\%. The types ``rel'' and ``obj'' are often misclassified as ``attr'' and ``cat'', respectively. This is because the questions are often short, making it difficult to distinguish between them.

To investigate how these classification errors affect the final QA accuracy, Table~\ref{tab:classification_ablation} compares AdaCoder using 1) predicted question types, 2) ground-truth question types, 3) random question types, and 4) without using question type based compression.
We observe three key findings.
First, the best performance is achieved by using ground truth question types. This highlights the importance of classifying question types to improve overall accuracy.
Second, the performance drop due to classification errors is less than 1.0 points.
This suggests that AdaCoder effectively classified the critical question types necessary for code generation, even though the accuracy for question classification is not very high.
Third, the method using random question types, which compresses prompts for each question type and randomly choose one of them in inference, is better than the method without question type based compression.
This is because the instruction prompt $\rsp$ in Eq.~(\ref{eq:compress_code_snippets}) for specializing code snippets to each question type makes it more likely to provide code snippets that are related to each other, thereby increasing the probability of completing the program. %In the case of simple compression,
When this instruction is omitted and compression is performed regardless of the question type, code snippets that are effective for any question type tend to be retained after compression.
However, this approach results in the loss of some specific snippets that are necessary to complete the program, thereby reducing QA accuracy. These results suggest that the instruction $\rsp$ is important for compressing code snippets.

\begin{table}[t]
\centering
\setlength{\tabcolsep}{3.2pt}
\caption{Analysis on effect of question type classification.}
\label{tab:classification_ablation}
\tabvspacemid
\begin{tabular}{l|c|c}
\toprule
Method & Token length & Accuracy (\%)\\
\midrule
w/ Predicted question types & 993 & 43.6 \\
w/ Oracle question types & 851 & 44.5 \\
w/ Random question types & 851 & 37.6\\
w/o Q. type based compression & 810 & 28.9\\
\bottomrule
\end{tabular}
\tabvspacebottom
\end{table}

\begin{figure}
\centering
\includegraphics[width=\linewidth]{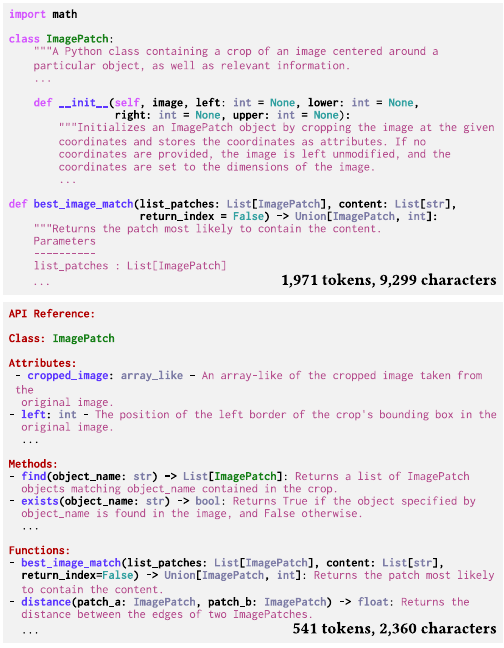}
\caption{Comparison of the original and compressed API definitions ($\pdef$ and $\pdefhat$). AdaCoder reduced the token length by 72.5\%.}
\label{fig:prompt_api_compress}
\end{figure}

\noindent \textbf{Compressed prompts.}
Table~\ref{tab:compression} summarizes the token length and compression performance for each component of the input preprompt.
We see that both API definitions and code snippets are significantly compressed.
A comparison of the original and compressed API definitions is shown in Figure~\ref{fig:prompt_api_compress}.
We see that descriptions of methods unnecessary for coding, such as those for the initialization method, are omitted, and the remaining sections are condensed into shorter sentences. This is an effective compression achieved by the language understanding and summarization capabilities of black-box LLMs.

\begin{table}[t]
\centering
\caption{Token length and number of characters for each component of input prompt. Reduction rate is measured by token length.}
\setlength{\tabcolsep}{1.5pt}
\tabvspacemid
\label{tab:compression}
\begin{tabular}{l|cc|ccc}
\toprule
\multirow{2}{*}{Component} & \multicolumn{2}{c|}{ViperGPT} & \multicolumn{3}{c}{AdaCoder}\\
& Tokens & Characters & Tokens & Characters & Reduction \\
\midrule
API defs & 1,971 & 9,299 & 541 & 2,360 & 72.6\%\\
Code snippets & 1,386 & 6,263 & 234 & 971 & 83.1\%\\
Instruction & 77 & 388 & 77 & 388 & - \\
Classification & 0&0&141&618 & - \\
\midrule
Total & 3,434 & 15,950 & 993 & 4,337 & 71.7\%\\
\bottomrule
\end{tabular}
\end{table}

\noindent \textbf{Computational time.}
Since the model weights and details of the black-box LLMs are not publicly available, and API response times can be affected by server congestion, a detailed analysis of computation times is not possible. However, the total time for experiments on the GQA dataset was reduced by 55\%.

\vspace{3pt}
\noindent \textbf{\large \ref{sec:res}.1 \hspace{4pt} Ablation study and analysis}
\vspace{3pt}

\noindent \textbf{Ablation study.}
Table~\ref{tab:ablation} presents the results of an ablation study. We see that both compression of API definitions and code snippets contribute to each other for both reducing the input token length and improving QA accuracy.
Table~\ref{tab:single} summarizes the QA accuracy obtained by using a single compressed prompt. We see that even with one prompt of either ``attr'' or ``rel'', our method achieves comparable or slightly better performance than the ViperGPT baseline (41.3\%). However, using one prompt of either ``obj'' or ``global'', the QA accuracy is significantly degraded. These results demonstrate that our adaptation approach is essential for improving QA accuracy while compressing input prompts.
The detailed QA accuracy by question type is analyzed in Table~\ref{tab:cross}.
We see that the four compressed prompt specialized for ``obj'', ``cat'', ``attr'', and ``rel'' performed the best for corresponding questions.
For the ``global'' questions, the prompt for ``attr'' was the best.
This is because ``global'' questions are highly varied and not easily categorized. 
Defining fine-grained QA types would be interesting as a next step in future research.
\begin{table}[t]
\setlength{\tabcolsep}{9pt}
\caption{Ablation study w.r.t prompt compression (GQA, gpt-3.5-turbo).}
\label{tab:ablation}
\tabvspacemid
\begin{tabular}{l|c|c}
\toprule
Method & Tokens & Acc.\\
\midrule
AdaCoder & 993 & \textbf{43.6} \\
\midrule
w/o compressing API defs. & 2,422 & 40.6\\
w/o compressing code snips. & 2,145 & 41.1\\
w/o QA classification & 810 & 28.9\\
w/o any compression & 3,434 & 41.3\\
\bottomrule
\end{tabular}
\tabvspacebottom
\end{table}
\begin{table}[t]
\setlength{\tabcolsep}{9pt}
\caption{Ablation study using a single specialized prompt (GQA, gpt-3.5-turbo).}
\label{tab:single}
\tabvspacemid
\begin{tabular}{l|c|c}
\toprule
Method & Tokens & Acc.\\
\midrule
AdaCoder & 993 & \textbf{43.6} \\
\midrule
w/ fixed prompt of obj & 967 & 30.9\\
w/ fixed prompt of cat & 1,015 & 39.0\\
w/ fixed prompt of attr & 1,008 & 41.7\\
w/ fixed prompt of rel & 993 & 42.3\\
w/ fixed prompt of global & 977 & 35.3\\
\bottomrule
\end{tabular}
\tabvspacebottom
\end{table}
\begin{table}[t]
\setlength{\tabcolsep}{8pt}
\caption{Cross question type evaluation (GQA, gpt-3.5-turbo).}
\label{tab:cross}
\tabvspacemid
\begin{tabular}{l|ccccc}
\toprule
QA type & obj & cat & attr & rel & global \\
\midrule
obj & \textbf{77.0} & 17.5 & 31.1 & 21.8 & 16.9 \\ 
cat & 74.5 & \textbf{45.3} & 39.9 & 28.5 & 35.4 \\
attr & 68.9 & 30.7 & \textbf{52.4} & 28.9 & \textbf{36.9} \\
rel & 70.2 & 35.8 & 50.9 & \textbf{30.3} & 33.9 \\
global & 71.1 & 31.4 & 38.0 & 24.8 & 32.3\\
\bottomrule
\end{tabular}
\tabvspacebottom
\end{table}
% moved to ablation
% moved to tab_ablation.tex
\begin{figure*}
\centering
\includegraphics[width=\linewidth]{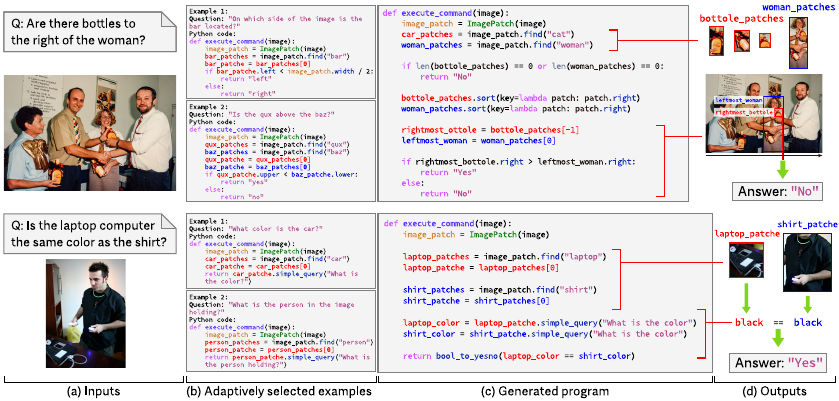}
\caption{Qualitative examples.
(a) Input of questions and images.
(b) Adaptively selected example code snippets. Each compressed prompt involves three or four snippets, and two of them are shown. These examples are fed into LLM with the compressed API definition in Figure~\ref{fig:prompt_api_compress}.
(c) Generated Python program for question answering.
(d) Visualization of intermediate outputs to derive the answer.
}
\label{fig:qualitative}
\end{figure*}

\noindent \textbf{Error analysis.}
Table~\ref{tab:errors} shows an error analysis, where we manually counted the occurrence of four types of errors.
``Coding error'' indicates that the generated program is not executable or returns nothing.
``Cannot answer to simple query'' indicates that the program is correct but the \texttt{simple\_query} method returned a response such as ``I cannot answer''.
``No object detected'' indicates that no object is detected by the \texttt{find} method.
``Wrong answer'' indicates that the returned answer was wrong.
We have two observations. First, the predominant type of error was wrong answers, and AdaCoder reduced their frequency.
Second, despite AdaCoder's improvement in coding quality, there is still a 7.8\% incidence of coding errors. This suggests that there is still room for improvement in instructing LLMs about API usage.

\begin{table}[t]
\centering
%\vspace{5.1pt}
\caption{Error analysis (individual error rates as percentages).}
\label{tab:errors}
\tabvspacemid
%\vspace{3.8pt}
\begin{tabular}{l|cc}
\toprule
Error type & ViperGPT & AdaCoder\\
\midrule
%Exact match & 41.3 & 43.6\\
%\midrule
Correct but with unnecessary details & 0.5 & 0.5\\
Correct except for articles & 1.1 & 1.5 \\
Correct by paraphrasing & 1.4 & 1.6\\
\midrule
Coding error & 8.3 & 7.8 \\
Cannot answer to simple query & 6.1 & 6.1\\
No object detected & 0.7 & 1.6\\
Wrong answer & 40.6 & 37.3\\
\bottomrule
\end{tabular}
\tabvspacebottom
\vspace{-8pt}
\end{table}

Several minor errors were also observed. ``Correct but with unnecessary details'' refers to responses that were marked incorrect because they provided additional, unnecessary information, such as the response ``Yes, there is an apple on the table'' where the ground truth is ``Yes''.
``Correct except with articles'' refers to cases where the instruction to respond with a single word was ignored and an article was added, resulting in responses such as ``a car'' instead of ``car''.
``Correct by paraphrasing'' refers to errors resulting from the use of interchangeable terms that do not change the meaning, such as using ``lady'' instead of ``woman''.

\noindent \textbf{Qualitative examples.}
Figure~\ref{fig:qualitative} presents qualitative examples of the generated programs.
As shown, few example code snippets related to the input question are adaptively selected.
These examples help LLM to generate a program to answer the question.
When the program is executed, the object patches are detected and then the relative position or colors are compared to derive a correct answer.

\noindent {\bf Evaluation on various tasks.}
Finally, we evaluated the generalizability of our method across various multimodal tasks, including visual grounding (VG), counting (CT), relative depth estimation (RDE), visual reasoning (VR), image editing (IE), and object tagging (OT).
As shown in Table~\ref{tab:various}, our method is consistently effective for all of these tasks.
\begin{table}[t]
\centering
\caption{Evaluation on various multimodal tasks.
Token and Red. indicate token length and reduction rate, respectively.
}
\setlength{\tabcolsep}{1pt}
\begin{tabular}{cl|c|cc|ccc}
\toprule
& \multirow{2}{*}{Task} & \multirow{2}{*}{Dataset} & \multicolumn{2}{c|}{Baseline}
& \multicolumn{3}{c}{AdaCoder}\\
& & & Acc. & Token & Acc. & Token & Red.$\uparrow$\\
\midrule
\multirow{3}{*}{\rotatebox{90}{\small ViperGPT}} & Visual grounding & RefCOCO \cite{yu2016modeling} & 39.5 & 3,434 & \textbf{49.0} & 853 & 75.2\% \\
& Counting & BLINK~\cite{fu2024blink} & 38.3 & 3,434 & \textbf{39.2} & 935 & 72.8\%\\
& Relative depth & BLINK~\cite{fu2024blink} & \textbf{59.7} & 3,434 & \textbf{59.7} & 847 & 75.3\% \\
\midrule
\multirow{4}{*}{\rotatebox{90}{\small VisProg}}
& Visual QA & GQA~\cite{Hudson2019GQADataset} & 47.5 & 1,836 & \textbf{49.0} & 657 & 64.2\%\\
& Visual reasoning & NLVR2~\cite{suhr2019nlvr2} & 62.0 & 1,495 & \textbf{62.5} & 338 & 77.4\%\\
& Image editing & MagicBrush~\cite{zhang2023magicbrush} & 68.3 & 879 & \textbf{70.0} & 219 & 75.1\%\\
& Object tagging & RefCOCO \cite{yu2016modeling} & \textbf{43.3} & 1,836 & \textbf{43.3} & 216 & 88.2\%\\
\bottomrule
\end{tabular}
\label{tab:various}
\end{table}

\section{Conclusion}
We introduced AdaCoder, a framework for adaptive prompt compression for visual programmatic models.
AdaCoder efficiently generated programs for visual question answering by compressing and selecting prompts depending on the question type.
A single black-box LLM is effectively employed to perform question type classification, textual compression and code generation, eliminating the need for additional training.
In experiments, we demonstrated the effectiveness and efficiency of AdaCoder in comparison to ViperGPT and LLMLingua.
Finally, we discuss limitations and future work.

\noindent \textbf{Limitations.}
As this work relies on black-box LLMs,
analysis from the perspective of neural network architecture is limited.
Alternative choices to LLMs for code generation may include open-source white-box models, such as CodeLlama and StarCoder. However, since AdaCoder requires high-quality text classification and summarization, these models were not suitable for prompt compression. 
New research directions leveraging the combination of white-box and black-box LLMs need to be further explored.

\noindent \textbf{Future work.}
To advance multimodal automated programming, future research directions that focus on pushing the boundaries beyond the traditional scope of VQA would be interesting. This includes developing methods for interactive code modification to enable a more dynamic and responsive programming environment.
Additionally, we plan to explore the automatic extension of APIs to facilitate their evolution in becoming more efficient and effective in addressing the complex requirements of multimodal interactions.
From a long-term perspective, we believe that adaptively extracting necessary API definitions and coding examples will be a critical step in automating large-scale multimodal software development to reduce computational costs and save energy.

\section*{Acknowledgements}
This work was supported by JSPS KAKENHI Grant Numbers 21H04910 and 22K12089.

\bibliographystyle{ACM-Reference-Format}
{
\bibliography{ref}


\begin{thebibliography}{48}

%%% the Web address.

\ifx \showCODEN    \undefined \def \showCODEN     #1{\unskip}     \fi
\ifx \showDOI      \undefined \def \showDOI       #1{#1}\fi
\ifx \showISBNx    \undefined \def \showISBNx     #1{\unskip}     \fi
\ifx \showISBNxiii \undefined \def \showISBNxiii  #1{\unskip}     \fi
\ifx \showISSN     \undefined \def \showISSN      #1{\unskip}     \fi
\ifx \showLCCN     \undefined \def \showLCCN      #1{\unskip}     \fi
\ifx \shownote     \undefined \def \shownote      #1{#1}          \fi
\ifx \showarticletitle \undefined \def \showarticletitle #1{#1}   \fi
\ifx \showURL      \undefined \def \showURL       {\relax}        \fi
% invisible to TeX
\providecommand\bibfield[2]{#2}
\providecommand\bibinfo[2]{#2}
\providecommand\natexlab[1]{#1}
\providecommand\showeprint[2][]{arXiv:#2}

\bibitem[Anderson et~al\mbox{.}(2018)]%
        {anderson2018bottomupandtopdown}
\bibfield{author}{\bibinfo{person}{Peter Anderson}, \bibinfo{person}{Xiaodong He}, \bibinfo{person}{Chris Buehler}, {et~al\mbox{.}}} \bibinfo{year}{2018}\natexlab{}.
\newblock \showarticletitle{Bottom-up and top-down attention for image captioning and visual question answering}. In \bibinfo{booktitle}{\emph{Proc. IEEE/CVF Conference on Computer Vision and Pattern Recognition (CVPR)}}. \bibinfo{pages}{6077--6086}.
\newblock


\bibitem[Anthropic Claude AIP.(2023)]%
        {claude}
Anthropic Claude AIP. \bibinfo{year}{2023}\natexlab{}.
\newblock
\newblock
\newblock
\shownote{https://claude.ai}.


\bibitem[Antol et~al\mbox{.}(2015)]%
        {antol2015vqa}
\bibfield{author}{\bibinfo{person}{Stanislaw Antol}, \bibinfo{person}{Aishwarya Agrawal}, \bibinfo{person}{Jiasen Lu}, \bibinfo{person}{Margaret Mitchell}, \bibinfo{person}{Dhruv Batra}, \bibinfo{person}{C~Lawrence Zitnick}, {and} \bibinfo{person}{Devi Parikh}.} \bibinfo{year}{2015}\natexlab{}.
\newblock \showarticletitle{{VQA}: Visual question answering}. In \bibinfo{booktitle}{\emph{Proc. IEEE/CVF International Conference on Computer Vision (ICCV)}}. \bibinfo{pages}{2425--2433}.
\newblock


\bibitem[Bolya et~al\mbox{.}(2023)]%
        {bolya23tokenmerging}
\bibfield{author}{\bibinfo{person}{Daniel Bolya}, \bibinfo{person}{Cheng-Yang Fu}, \bibinfo{person}{Xiaoliang Dai}, \bibinfo{person}{Peizhao Zhang}, \bibinfo{person}{Christoph Feichtenhofer}, {and} \bibinfo{person}{Judy Hoffman}.} \bibinfo{year}{2023}\natexlab{}.
\newblock \showarticletitle{Token Merging: Your ViT But Faster}. In \bibinfo{booktitle}{\emph{Proc. International Conference on Learning Representations (ICLR)}}.
\newblock


\bibitem[Brown et~al\mbox{.}(2020)]%
        {brown2020gpt3}
\bibfield{author}{\bibinfo{person}{Tom Brown}, \bibinfo{person}{Benjamin Mann}, \bibinfo{person}{Nick Ryder}, {et~al\mbox{.}}} \bibinfo{year}{2020}\natexlab{}.
\newblock \showarticletitle{Language models are few-shot learners}. In \bibinfo{booktitle}{\emph{Proc. Annual Conference on Neural Information Processing Systems (NeurIPS)}}. \bibinfo{pages}{1877--1901}.
\newblock


\bibitem[Chen et~al\mbox{.}(2023a)]%
        {chen23kecvqg}
\bibfield{author}{\bibinfo{person}{Jiali Chen}, \bibinfo{person}{Zhenjun Guo}, \bibinfo{person}{Jiayuan Xie}, \bibinfo{person}{Yi Cai}, {and} \bibinfo{person}{Qing Li}.} \bibinfo{year}{2023}\natexlab{a}.
\newblock \showarticletitle{Deconfounded Visual Question Generation with Causal Inference}. In \bibinfo{booktitle}{\emph{Proc. ACM International Conference on Multimedia (ACMMM)}}. \bibinfo{pages}{5132--5142}.
\newblock


\bibitem[Chen et~al\mbox{.}(2023b)]%
        {chen2023vtqa2023}
\bibfield{author}{\bibinfo{person}{Kang Chen}, \bibinfo{person}{Tianli Zhao}, {and} \bibinfo{person}{Xiangqian Wu}.} \bibinfo{year}{2023}\natexlab{b}.
\newblock \showarticletitle{VTQA2023: ACM Multimedia 2023 Visual Text Question Answering Challenge}. In \bibinfo{booktitle}{\emph{Proc. ACM International Conference on Multimedia (ACMMM)}}. \bibinfo{pages}{9646--9650}.
\newblock


\bibitem[Chen et~al\mbox{.}(2021)]%
        {Chen2021Codex}
\bibfield{author}{\bibinfo{person}{Mark Chen}, \bibinfo{person}{Jerry Tworek}, {et~al\mbox{.}}} \bibinfo{year}{2021}\natexlab{}.
\newblock \showarticletitle{Evaluating Large Language Models Trained on Code}.
\newblock \bibinfo{journal}{\emph{arXiv2107.03374}} (\bibinfo{year}{2021}).
\newblock


\bibitem[Chevalier et~al\mbox{.}(2023)]%
        {chevalier2023adapting}
\bibfield{author}{\bibinfo{person}{Alexis Chevalier}, \bibinfo{person}{Alexander Wettig}, \bibinfo{person}{Anirudh Ajith}, {and} \bibinfo{person}{Danqi Chen}.} \bibinfo{year}{2023}\natexlab{}.
\newblock \showarticletitle{Adapting Language Models to Compress Contexts}. In \bibinfo{booktitle}{\emph{Proc. Conference on Empirical Methods in Natural Language Processing (EMNLP)}}.
\newblock


\bibitem[Fu et~al\mbox{.}(2024)]%
        {fu2024blink}
\bibfield{author}{\bibinfo{person}{Xiang Fu}, \bibinfo{person}{Yuxin Hu}, \bibinfo{person}{Baoxu Li}, \bibinfo{person}{Yasheng Feng}, \bibinfo{person}{Hao Wang}, \bibinfo{person}{Xian Lin}, \bibinfo{person}{Dan Roth}, \bibinfo{person}{Noah~A. Smith}, \bibinfo{person}{Wei-Chiu Ma}, {and} \bibinfo{person}{Ranjay Krishna}.} \bibinfo{year}{2024}\natexlab{}.
\newblock \showarticletitle{BLINK: Multimodal Large Language Models Can See but Not Perceive}. In \bibinfo{booktitle}{\emph{Proc. European Conference on Computer Vision (ECCV)}}.
\newblock


\bibitem[Ge et~al\mbox{.}(2024)]%
        {ge2024recursive}
\bibfield{author}{\bibinfo{person}{Jiaxin Ge}, \bibinfo{person}{Sanjay Subramanian}, \bibinfo{person}{Baifeng Shi}, \bibinfo{person}{Roei Herzig}, {and} \bibinfo{person}{Trevor Darrell}.} \bibinfo{year}{2024}\natexlab{}.
\newblock \showarticletitle{Recursive Visual Programming}. In \bibinfo{booktitle}{\emph{Proc. European Conference on Computer Vision (ECCV)}}.
\newblock


\bibitem[Ge et~al\mbox{.}(2022)]%
        {ge2022extensible}
\bibfield{author}{\bibinfo{person}{Tao Ge}, \bibinfo{person}{Jing Hu}, \bibinfo{person}{Li Dong}, \bibinfo{person}{Shaoguang Mao}, \bibinfo{person}{Yan Xia}, \bibinfo{person}{Xun Wang}, \bibinfo{person}{Si-Qing Chen}, {and} \bibinfo{person}{Furu Wei}.} \bibinfo{year}{2022}\natexlab{}.
\newblock \showarticletitle{Extensible Prompts for Language Models on Zero-shot Language Style Customization}. In \bibinfo{booktitle}{\emph{Proc. Annual Conference on Neural Information Processing Systems (NeurIPS)}}.
\newblock


\bibitem[Goyal et~al\mbox{.}(2020)]%
        {goyal20powerbert}
\bibfield{author}{\bibinfo{person}{Saurabh Goyal}, \bibinfo{person}{Anamitra~Roy Choudhury}, \bibinfo{person}{Saurabh Raje}, \bibinfo{person}{Venkatesan~T. Chakaravarthy}, \bibinfo{person}{Yogish Sabharwal}, {and} \bibinfo{person}{Ashish Verma}.} \bibinfo{year}{2020}\natexlab{}.
\newblock \showarticletitle{Power-bert: Accelerating BERT inference via progressive word-vector elimination}. In \bibinfo{booktitle}{\emph{Proc. International Conference on Machine Learning (ICML)}}. \bibinfo{pages}{3690--3699}.
\newblock


\bibitem[Goyal et~al\mbox{.}(2017)]%
        {goyal2017vqav2}
\bibfield{author}{\bibinfo{person}{Yash Goyal}, \bibinfo{person}{Tejas Khot}, {et~al\mbox{.}}} \bibinfo{year}{2017}\natexlab{}.
\newblock \showarticletitle{Making the V in VQA matter: Elevating the role of image understanding in Visual Question Answering}. In \bibinfo{booktitle}{\emph{Proc. IEEE/CVF Conference on Computer Vision and Pattern Recognition (CVPR)}}. \bibinfo{pages}{6904--6913}.
\newblock


\bibitem[Gupta and Kembhavi(2022)]%
        {Gupta2022VisProg}
\bibfield{author}{\bibinfo{person}{Tanmay Gupta} {and} \bibinfo{person}{Aniruddha Kembhavi}.} \bibinfo{year}{2022}\natexlab{}.
\newblock \showarticletitle{Visual Programming: Compositional visual reasoning without training}. In \bibinfo{booktitle}{\emph{Proc. IEEE/CVF Conference on Computer Vision and Pattern Recognition (CVPR)}}.
\newblock


\bibitem[Hu et~al\mbox{.}(2024)]%
        {Hu2024VPD}
\bibfield{author}{\bibinfo{person}{Yushi Hu}, \bibinfo{person}{Otilia Stretcu}, \bibinfo{person}{Chun-Ta Lu}, \bibinfo{person}{Krishnamurthy Viswanathan}, \bibinfo{person}{Kenji Hata}, \bibinfo{person}{Enming Luo}, \bibinfo{person}{Ranjay Krishna}, {and} \bibinfo{person}{Ariel Fuxman}.} \bibinfo{year}{2024}\natexlab{}.
\newblock \showarticletitle{Visual Program Distillation: Distilling Tools and Programmatic Reasoning into Vision-Language Models}. In \bibinfo{booktitle}{\emph{Proc. IEEE/CVF Conference on Computer Vision and Pattern Recognition (CVPR)}}. \bibinfo{pages}{9590--9601}.
\newblock


\bibitem[Huang et~al\mbox{.}(2021)]%
        {huang2021diagnostic}
\bibfield{author}{\bibinfo{person}{Ziqi Huang}, \bibinfo{person}{Hongyuan Zhu}, \bibinfo{person}{Ying Sun}, {et~al\mbox{.}}} \bibinfo{year}{2021}\natexlab{}.
\newblock \showarticletitle{A diagnostic study of visual question answering with analogical reasoning}. In \bibinfo{booktitle}{\emph{Proc. IEEE International Conference on Image Processing (ICIP)}}. IEEE, \bibinfo{pages}{2463--2467}.
\newblock


\bibitem[Hudson and Manning(2019)]%
        {Hudson2019GQADataset}
\bibfield{author}{\bibinfo{person}{Drew~A. Hudson} {and} \bibinfo{person}{Christopher~D. Manning}.} \bibinfo{year}{2019}\natexlab{}.
\newblock \showarticletitle{GQA: A New Dataset for Real-World Visual Reasoning and Compositional Question Answering}. In \bibinfo{booktitle}{\emph{Proc. IEEE/CVF International Conference on Computer Vision (ICCV)}}.
\newblock


\bibitem[Jiang et~al\mbox{.}(2023)]%
        {jiang2023llmlingua}
\bibfield{author}{\bibinfo{person}{Huiqiang Jiang}, \bibinfo{person}{Qianhui Wu}, \bibinfo{person}{Chin-Yew Lin}, \bibinfo{person}{Yuqing Yang}, {and} \bibinfo{person}{Lili Qiu}.} \bibinfo{year}{2023}\natexlab{}.
\newblock \showarticletitle{{LLML}ingua: Compressing Prompts for Accelerated Inference of Large Language Models}. In \bibinfo{booktitle}{\emph{Proc. Conference on Empirical Methods in Natural Language Processing (EMNLP)}}.
\newblock


\bibitem[Jin and Lu(2023)]%
        {Jin2023TabularCoT}
\bibfield{author}{\bibinfo{person}{Ziqi Jin} {and} \bibinfo{person}{Wei Lu}.} \bibinfo{year}{2023}\natexlab{}.
\newblock \showarticletitle{Tab-CoT: Zero-shot Tabular Chain of Thought}. In \bibinfo{booktitle}{\emph{Proc. Findings of the Association for Computational Linguistics (ACL Findings)}}.
\newblock


\bibitem[Kim and Cho(2021)]%
        {kim21lengthadaptive}
\bibfield{author}{\bibinfo{person}{Gyuwan Kim} {and} \bibinfo{person}{Kyunghyun Cho}.} \bibinfo{year}{2021}\natexlab{}.
\newblock \showarticletitle{Length-adaptive transformer: Train once with length drop, use anytime with search}. In \bibinfo{booktitle}{\emph{Proc. Annual Meeting of the Association for Computational Linguistics (ACL)}}. \bibinfo{pages}{6501--6511}.
\newblock


\bibitem[Kim et~al\mbox{.}(2022)]%
        {kim22learned}
\bibfield{author}{\bibinfo{person}{Sehoon Kim}, \bibinfo{person}{Sheng Shen}, \bibinfo{person}{David Thorsley}, \bibinfo{person}{Amir Gholami}, \bibinfo{person}{Woosuk Kwon}, \bibinfo{person}{Joseph Hassoun}, {and} \bibinfo{person}{Kurt Keutzer}.} \bibinfo{year}{2022}\natexlab{}.
\newblock \showarticletitle{Learned Token Pruning for Transformers}. In \bibinfo{booktitle}{\emph{Proc. ACM SIGKDD Conference on Knowledge Discovery and Data Mining (SIGKDD)}}. \bibinfo{pages}{784--794}.
\newblock


\bibitem[Lan et~al\mbox{.}(2023)]%
        {lan23questionprompts}
\bibfield{author}{\bibinfo{person}{Yunshi Lan}, \bibinfo{person}{Xiang Li}, \bibinfo{person}{Xin Liu}, \bibinfo{person}{Yang Li}, \bibinfo{person}{Wei Qin}, {and} \bibinfo{person}{Weining Qian}.} \bibinfo{year}{2023}\natexlab{}.
\newblock \showarticletitle{Improving Zero-shot Visual Question Answering via Large Language Models with Reasoning Question Prompts}. In \bibinfo{booktitle}{\emph{Proc. ACM International Conference on Multimedia (ACMMM)}}.
\newblock


\bibitem[Le et~al\mbox{.}(2021)]%
        {le2021vttransformer}
\bibfield{author}{\bibinfo{person}{Tung Le}, \bibinfo{person}{Huy~Tien Nguyen}, {and} \bibinfo{person}{Minh Le~Nguyen}.} \bibinfo{year}{2021}\natexlab{}.
\newblock \showarticletitle{Vision and text transformer for predicting answerability on visual question answering}. In \bibinfo{booktitle}{\emph{Proc. IEEE International Conference on Image Processing (ICIP)}}. IEEE, \bibinfo{pages}{934--938}.
\newblock


\bibitem[Li et~al\mbox{.}(2021)]%
        {li2021unsupervised}
\bibfield{author}{\bibinfo{person}{Liunian~Harold Li}, \bibinfo{person}{Haoxuan You}, \bibinfo{person}{Zhecan Wang}, {et~al\mbox{.}}} \bibinfo{year}{2021}\natexlab{}.
\newblock \showarticletitle{Unsupervised Vision-and-Language Pre-training Without Parallel Images and Captions}. In \bibinfo{booktitle}{\emph{Proc. Annual Conference of the North American Chapter of the Association for Computational Linguistics (NAACL)}}.
\newblock


\bibitem[Li et~al\mbox{.}(2022b)]%
        {Li2022GLIP}
\bibfield{author}{\bibinfo{person}{Liunian~Harold Li}, \bibinfo{person}{Pengchuan Zhang}, \bibinfo{person}{Haotian Zhang}, \bibinfo{person}{Jianwei Yang}, \bibinfo{person}{Chunyuan Li}, \bibinfo{person}{Yiwu Zhong}, \bibinfo{person}{Lijuan Wang}, \bibinfo{person}{Lu Yuan}, \bibinfo{person}{Lei Zhang}, \bibinfo{person}{Jenq-Neng Hwang}, {et~al\mbox{.}}} \bibinfo{year}{2022}\natexlab{b}.
\newblock \showarticletitle{Grounded language-image pre-training}. In \bibinfo{booktitle}{\emph{Proc. IEEE/CVF Conference on Computer Vision and Pattern Recognition (CVPR)}}.
\newblock


\bibitem[Li et~al\mbox{.}(2022a)]%
        {Li22aivqa}
\bibfield{author}{\bibinfo{person}{Rengang Li}, \bibinfo{person}{Cong Xu}, \bibinfo{person}{Zhenhua Guo}, \bibinfo{person}{Baoyu Fan}, \bibinfo{person}{Runze Zhang}, \bibinfo{person}{Wei Liu}, \bibinfo{person}{Yaqian Zhao}, \bibinfo{person}{Weifeng Gong}, {and} \bibinfo{person}{Endong Wang}.} \bibinfo{year}{2022}\natexlab{a}.
\newblock \showarticletitle{AI-VQA: Visual Question Answering based on Agent Interaction with Interpretability}. In \bibinfo{booktitle}{\emph{Proc. ACM International Conference on Multimedia (ACMMM)}}. \bibinfo{pages}{5274--5282}.
\newblock


\bibitem[Malinowski et~al\mbox{.}(2015)]%
        {malinowski2015ask}
\bibfield{author}{\bibinfo{person}{Mateusz Malinowski}, \bibinfo{person}{Marcus Rohrbach}, {and} \bibinfo{person}{Mario Fritz}.} \bibinfo{year}{2015}\natexlab{}.
\newblock \showarticletitle{Ask Your Neurons: A Neural-based Approach to Answering Questions about Images}. In \bibinfo{booktitle}{\emph{Proc. IEEE/CVF International Conference on Computer Vision (ICCV)}}. \bibinfo{pages}{1--9}.
\newblock


\bibitem[Mu et~al\mbox{.}(2023)]%
        {mu2023learning}
\bibfield{author}{\bibinfo{person}{Jesse Mu}, \bibinfo{person}{Xiang~Lisa Li}, {and} \bibinfo{person}{Noah Goodman}.} \bibinfo{year}{2023}\natexlab{}.
\newblock \showarticletitle{Learning to Compress Prompts with Gist Tokens}. In \bibinfo{booktitle}{\emph{Proc. Annual Conference on Neural Information Processing Systems (NeurIPS)}}.
\newblock


\bibitem[Pan et~al\mbox{.}(2024)]%
        {wu2024llmlingua2}
\bibfield{author}{\bibinfo{person}{Zhuoshi Pan}, \bibinfo{person}{Qianhui Wu}, \bibinfo{person}{Huiqiang Jiang}, \bibinfo{person}{Menglin Xia}, \bibinfo{person}{Xufang Luo}, \bibinfo{person}{Jue Zhang}, \bibinfo{person}{Qingwei Lin}, \bibinfo{person}{Victor R{\"u}hle}, \bibinfo{person}{Yuqing Yang}, \bibinfo{person}{Chin-Yew Lin}, \bibinfo{person}{H.~Vicky Zhao}, \bibinfo{person}{Lili Qiu}, {and} \bibinfo{person}{Dongmei Zhang}.} \bibinfo{year}{2024}\natexlab{}.
\newblock \showarticletitle{LLMLingua-2: Data Distillation for Efficient and Faithful Task-Agnostic Prompt Compression}.
\newblock \bibinfo{journal}{\emph{arXiv preprint arXiv:2403.12968}} (\bibinfo{year}{2024}).
\newblock


\bibitem[Roziere et~al\mbox{.}(2023)]%
        {roziere2023codellama}
\bibfield{author}{\bibinfo{person}{Baptiste Roziere}, \bibinfo{person}{Jonas Gehring}, \bibinfo{person}{Fabian Gloeckle}, {et~al\mbox{.}}} \bibinfo{year}{2023}\natexlab{}.
\newblock \showarticletitle{Code llama: Open foundation models for code}.
\newblock \bibinfo{journal}{\emph{arXiv preprint arXiv:2308.12950}} (\bibinfo{year}{2023}).
\newblock


\bibitem[Sarkar and Rahnemoonfar(2022)]%
        {sarkar2022grad}
\bibfield{author}{\bibinfo{person}{Argho Sarkar} {and} \bibinfo{person}{Maryam Rahnemoonfar}.} \bibinfo{year}{2022}\natexlab{}.
\newblock \showarticletitle{{Grad-CAM} aware supervised attention for visual question answering for post-disaster damage assessment}. In \bibinfo{booktitle}{\emph{Proc. IEEE International Conference on Image Processing (ICIP)}}. IEEE, \bibinfo{pages}{3783--3787}.
\newblock


\bibitem[Shen et~al\mbox{.}(2024)]%
        {shen2024pyramid}
\bibfield{author}{\bibinfo{person}{Ruoyue Shen}, \bibinfo{person}{Nakamasa Inoue}, {and} \bibinfo{person}{Koichi Shinoda}.} \bibinfo{year}{2024}\natexlab{}.
\newblock \showarticletitle{Pyramid Coder: Hierarchical Code Generator for Compositional Visual Question Answering}. In \bibinfo{booktitle}{\emph{Proc. IEEE International Conference on Image Processing (ICIP)}}.
\newblock


\bibitem[Subramanian et~al\mbox{.}(2023)]%
        {Subramanian2023CodeVQA}
\bibfield{author}{\bibinfo{person}{Sanjay Subramanian}, \bibinfo{person}{Medhini Narasimhan}, {et~al\mbox{.}}} \bibinfo{year}{2023}\natexlab{}.
\newblock \showarticletitle{Modular Visual Question Answering via Code Generation}. In \bibinfo{booktitle}{\emph{Proc. Annual Meeting of the Association for Computational Linguistics (ACL)}}.
\newblock


\bibitem[Suhr et~al\mbox{.}(2019)]%
        {suhr2019nlvr2}
\bibfield{author}{\bibinfo{person}{Alane Suhr}, \bibinfo{person}{Stephanie Zhou}, \bibinfo{person}{Ally Zhang}, \bibinfo{person}{Iris Zhang}, \bibinfo{person}{Huajun Bai}, {and} \bibinfo{person}{Yoav Artzi}.} \bibinfo{year}{2019}\natexlab{}.
\newblock \showarticletitle{A Corpus for Reasoning About Natural Language Grounded in Photographs}. In \bibinfo{booktitle}{\emph{Proc. Annual Meeting of the Association for Computational Linguistics (ACL)}}. \bibinfo{pages}{6418--6428}.
\newblock


\bibitem[Sur\'is et~al\mbox{.}(2023)]%
        {Suris2023ViperGPT}
\bibfield{author}{\bibinfo{person}{D\'idac Sur\'is}, \bibinfo{person}{Sachit Menon}, {and} \bibinfo{person}{Carl Vondrick}.} \bibinfo{year}{2023}\natexlab{}.
\newblock \showarticletitle{{ViperGPT}: Visual Inference via Python Execution for Reasoning}. In \bibinfo{booktitle}{\emph{Proc. IEEE/CVF International Conference on Computer Vision (ICCV)}}.
\newblock


\bibitem[Team et~al\mbox{.}(2023)]%
        {team2023gemini}
\bibfield{author}{\bibinfo{person}{Gemini Team}, \bibinfo{person}{Rohan Anil}, {et~al\mbox{.}}} \bibinfo{year}{2023}\natexlab{}.
\newblock \showarticletitle{Gemini: a family of highly capable multimodal models}.
\newblock \bibinfo{journal}{\emph{arXiv preprint arXiv:2312.11805}} (\bibinfo{year}{2023}).
\newblock


\bibitem[Touvron et~al\mbox{.}(2023)]%
        {touvron2023llama}
\bibfield{author}{\bibinfo{person}{Hugo Touvron}, \bibinfo{person}{Louis Martin}, \bibinfo{person}{Kevin Stone}, {et~al\mbox{.}}} \bibinfo{year}{2023}\natexlab{}.
\newblock \showarticletitle{Llama 2: Open foundation and fine-tuned chat models}.
\newblock \bibinfo{journal}{\emph{arXiv preprint arXiv:2307.09288}} (\bibinfo{year}{2023}).
\newblock


\bibitem[Wang et~al\mbox{.}(2022)]%
        {wang22towards}
\bibfield{author}{\bibinfo{person}{Qingqing Wang}, \bibinfo{person}{Liqiang Xiao}, \bibinfo{person}{Yue Lu}, \bibinfo{person}{Yaohui Jin}, {and} \bibinfo{person}{Hao He}.} \bibinfo{year}{2022}\natexlab{}.
\newblock \showarticletitle{Towards Reasoning Ability in Scene Text Visual Question Answering}. In \bibinfo{booktitle}{\emph{Proc. ACM International Conference on Multimedia (ACMMM)}}. \bibinfo{pages}{2281--2289}.
\newblock


\bibitem[Wingate et~al\mbox{.}(2022)]%
        {wingate2022prompt}
\bibfield{author}{\bibinfo{person}{David Wingate}, \bibinfo{person}{Mohammad Shoeybi}, {and} \bibinfo{person}{Taylor Sorensen}.} \bibinfo{year}{2022}\natexlab{}.
\newblock \showarticletitle{Prompt Compression and Contrastive Conditioning for Controllability and Toxicity Reduction in Language Models}. In \bibinfo{booktitle}{\emph{Proc. Findings of Empirical Methods in Natural Language Processing (EMNLP Findings)}}.
\newblock


\bibitem[Wu et~al\mbox{.}(2022)]%
        {wu2022ques}
\bibfield{author}{\bibinfo{person}{Xiangyu Wu}, \bibinfo{person}{Jianfeng Lu}, \bibinfo{person}{Zhuanfeng Li}, {and} \bibinfo{person}{Fengchao Xiong}.} \bibinfo{year}{2022}\natexlab{}.
\newblock \showarticletitle{Ques-to-Visual Guided Visual Question Answering}. In \bibinfo{booktitle}{\emph{Proc. IEEE International Conference on Image Processing (ICIP)}}. IEEE, \bibinfo{pages}{4193--4197}.
\newblock


\bibitem[Yang et~al\mbox{.}(2022)]%
        {yang2022avqa}
\bibfield{author}{\bibinfo{person}{Pinci Yang}, \bibinfo{person}{Xin Wang}, \bibinfo{person}{Xuguang Duan}, \bibinfo{person}{Hong Chen}, \bibinfo{person}{Runze Hou}, \bibinfo{person}{Cong Jin}, {and} \bibinfo{person}{Wenwu Zhu}.} \bibinfo{year}{2022}\natexlab{}.
\newblock \showarticletitle{AVQA: A Dataset for Audio-Visual Question Answering on Videos}. In \bibinfo{booktitle}{\emph{Proc. ACM International Conference on Multimedia (ACMMM)}}. \bibinfo{pages}{3480--3491}.
\newblock


\bibitem[Yang et~al\mbox{.}(2016)]%
        {yang2016stackedattention}
\bibfield{author}{\bibinfo{person}{Zichao Yang}, \bibinfo{person}{Xiaodong He}, \bibinfo{person}{Jianfeng Gao}, {et~al\mbox{.}}} \bibinfo{year}{2016}\natexlab{}.
\newblock \showarticletitle{Stacked Attention Networks for Image Question Answering}. In \bibinfo{booktitle}{\emph{Proc. IEEE/CVF Conference on Computer Vision and Pattern Recognition (CVPR)}}.
\newblock


\bibitem[Yao et~al\mbox{.}(2023)]%
        {Yao2023TreeOfThoughts_ToT}
\bibfield{author}{\bibinfo{person}{Shunyu Yao}, \bibinfo{person}{Dian Yu}, \bibinfo{person}{Jeffrey Zhao}, {et~al\mbox{.}}} \bibinfo{year}{2023}\natexlab{}.
\newblock \showarticletitle{{Tree of Thoughts}: Deliberate Problem Solving with Large Language Models}. In \bibinfo{booktitle}{\emph{Proc. Annual Conference on Neural Information Processing Systems (NeurIPS)}}.
\newblock


\bibitem[Yu et~al\mbox{.}(2016)]%
        {yu2016modeling}
\bibfield{author}{\bibinfo{person}{Licheng Yu}, \bibinfo{person}{Patrick Poirson}, \bibinfo{person}{Shan Yang}, \bibinfo{person}{Alexander~C. Berg}, {and} \bibinfo{person}{Tamara~L. Berg}.} \bibinfo{year}{2016}\natexlab{}.
\newblock \showarticletitle{Modeling context in referring expressions}. In \bibinfo{booktitle}{\emph{Proc. European Conference on Computer Vision (ECCV)}}. \bibinfo{pages}{69--85}.
\newblock


\bibitem[Yuan et~al\mbox{.}(2023)]%
        {yuan23selfpt}
\bibfield{author}{\bibinfo{person}{Bowen Yuan}, \bibinfo{person}{Sisi You}, {and} \bibinfo{person}{Bing-Kun Bao}.} \bibinfo{year}{2023}\natexlab{}.
\newblock \showarticletitle{Self-PT: Adaptive Self-Prompt Tuning for Low-Resource Visual Question Answering}. In \bibinfo{booktitle}{\emph{Proc. ACM International Conference on Multimedia (ACMMM)}}. \bibinfo{pages}{5089--5098}.
\newblock


\bibitem[Zhang and Wu(2022)]%
        {zhang2022context}
\bibfield{author}{\bibinfo{person}{Haotian Zhang} {and} \bibinfo{person}{Wei Wu}.} \bibinfo{year}{2022}\natexlab{}.
\newblock \showarticletitle{Context Relation Fusion Model for Visual Question Answering}. In \bibinfo{booktitle}{\emph{Proc. IEEE International Conference on Image Processing (ICIP)}}. IEEE, \bibinfo{pages}{2112--2116}.
\newblock


\bibitem[Zhang et~al\mbox{.}(2023)]%
        {zhang2023magicbrush}
\bibfield{author}{\bibinfo{person}{Kai Zhang}, \bibinfo{person}{Li Mo}, \bibinfo{person}{Wei Chen}, \bibinfo{person}{Hao Sun}, {and} \bibinfo{person}{Yejin Su}.} \bibinfo{year}{2023}\natexlab{}.
\newblock \showarticletitle{MagicBrush: A manually annotated dataset for instruction-guided image editing}. In \bibinfo{booktitle}{\emph{Proc. Annual Conference on Neural Information Processing Systems (NeurIPS)}}.
\newblock


\end{thebibliography}
}
\end{document}